\documentclass[wcp]{jmlr}
\usepackage{url}
\usepackage{epstopdf}
\usepackage{slashbox}
\DeclareGraphicsExtensions{.pdf,.eps,.png,.jpg,.mps} 
\usepackage{colortbl}
\definecolor{maroon}{cmyk}{0,0.87,0.68,0.32}

\jmlrvolume{1}
\jmlryear{2016}
\jmlrworkshop{ICML 2016 AutoML Workshop}

\title[Average Ranking]{Effect of Incomplete Meta-dataset on Average Ranking Method}

  \author{\Name{Salisu Mamman Abdulrahman} \Email{salisu.abdul@gmail.com} \\
   \addr LIAAD - INESC TEC/Faculdade de Ci\^encias da Universidade do Porto \\
  \AND 
   \Name{Pavel Brazdil} \Email{pbrazdil@inescporto.pt}\\
   \addr LIAAD - INESC TEC/Faculdade de Economia,  Universidade do Porto\\}

\begin{document}

\maketitle

\begin{abstract}
One of the simplest metalearning methods is the average ranking method. 
This method uses metadata in the form of test results of a given set of algorithms on given set of datasets and calculates an average rank for each algorithm. 
The ranks are used to construct the average ranking. 
We investigate the problem of how the process of generating the average ranking is affected by incomplete metadata including fewer test results. 
This issue is relevant, because if we could show that incomplete metadata does not affect the final results much, we could explore it in future design. 
We could simply conduct fewer tests and save thus computation time. 
In this paper\footnote{This paper is a slightly updated version of the paper presented at AutoML workshop at ICML 2016, New York} we describe an upgraded average ranking method that is capable of dealing with incomplete metadata. 
Our results show that the proposed method is relatively robust to omission in test results in the meta datasets.
 
\end{abstract}
\begin{keywords}
{Average Ranking, Aggregation of Rankings, Incomplete Metadata}
\end{keywords}

\section{Introduction}
A large number of data mining algorithms exist, rooted in the fields of machine learning, statistics, pattern recognition, artificial intelligence. The task to recommend the most suitable algorithms has thus become rather challenging. 
The algorithm selection problem, originally described by \citet{Rice1976}, has attracted a great deal of attention, as it endeavours to select and apply the best or near best algorithm(s) for a given task (\citet{Brazdil2008,Smith2008}).
We address the problem of robustness of one particular version of an average ranking method that uses incomplete rankings as input. 
These arise if we have incomplete test results in the meta-dataset. 
We have investigated how much the performance degrades under such circumstances. 

The remainder of this paper is organized as follows. 
In the next section we present an overview of existing work in related areas. 
Section~\ref{sec:effect-of-incomplete} provide details about the proposed aggregation method for incomplete meta dataset, the experimental results and future work.

\section{Related Work}
\label{sec:relatedwork}
In this paper we are addressing a particular case of the algorithm selection problem, oriented towards the selection of classification algorithms. Various researchers addressed this problem in the course of the last 25 years. 
One approach to algorithm selection/recommendation relies on metalearning. 
The simplest method uses just performance results 
on different datasets in the form of rankings.
The rankings are then aggregated to obtain a single aggregated ranking.
So the aggregated ranking can be used as a simple model
that can be followed by the user to test the top candidates
to identify the algorithm to be used.
This strategy is sometimes referred to as the $Top$-$N$ strategy~(\citet{Brazdil2008}).

A more advanced approach often considered as the \emph{classical metalearning approach}
uses, in addition to performance results, also a set of measures that characterize datasets 
(\citet{Pfahringer2000,Brazdil2008,Smith2008}). 
However, this line is not followed up here.

Aggregation of rankings involving \emph{complete rankings} is a simple matter. 
Normally it just involves calculating the average rank for all items in the ranking
\citep{Lin2010}. 
Complete rankings are those in which $k$ items are ranked $N$ times and 
no value in this set is missing. 
Incomplete rankings arise when only some ranks are known in some of the rankings. 
These arise quite often in practice. 
Many diverse methods exist for aggregation of incomplete rankings. According to \citet{Lin2010} methods applicable to long lists can be divided into three categories: Heuristic algorithms, Markov chain methods and stochastic optimization methods.
The last category includes, for instance, \emph{Cross Entropy Monte Carlo, CEMC} method.

Some of the approaches require that the elements that do not appear 
in list $L_i$ of $k$ elements be attributed a concrete rank (e.g. $k+1$).  
This does not seem to be correct. 
We should not be forced to assume that some information exists, if in fact we have none.
We have considered using a package of R \emph{RankAggreg} (\citet{Pihur2014}),
but unfortunately we would have to attribute a concrete rank (e.g. $k+1$) 
to all missing elements. 
We have therefore developed a simple heuristic method based on Borda's method 
described in \citet{Lin2010}.
In our view it serves well our purpose.
As \citet{Lin2010} pointed out simple methods often compete quite well 
with other more complex approaches.
\section{Effect of Incomplete Meta-dataset on Average Ranking Method}
\label{sec:effect-of-incomplete}  
  
Our aim is to investigate the issue of how the generation of the average ranking is affected by incomplete test results in the meta-dataset available.
Here we focus on rankings obtained on the basis of accuracies. 
% although other measures, such as AUC, could have been used instead. 
%
%The term \emph{ranking} is used to indicate a ranking 
%obtained on the basis of a combined measure of accuracy and time (\citet{Abdulrahman2014}):
%$\mathit{A3R}_{a_\mathit{ref},a_j}^{d_i}=
%(\mathit{SR}_{a_j}^{d_i} /
%{\mathit{SR}_{a_\mathit{ref}}^{d_i}})/
%(\sqrt[N]{T_{a_j}^{d_i} / T_{a_\mathit{ref}}^{d_i}}
%$.
%
We wish to see how robust the method is to omissions in the meta-dataset. 
This issue is relevant because 
the meta-dataset that has been gathered by researchers is very often incomplete. 
Here we consider two different ways in which the meta-dataset can be incomplete:
First, the test results on some datasets may be completely missing. 
Second, there may be certain proportion of omissions in the test results of some algorithms on each dataset.

The expectation is that the performance of the average ranking method would degrade when less information is available. 
However, an interesting question is how grave the degradation is.
The answer to this issue is not straightforward, as it depends greatly on how diverse the datasets are and how this affects the rankings of algorithms.   
If the rankings are very similar, then we expect that the omissions would not make much difference. 
So the issue of the effects of omissions needs to be relativized.
To do this we will investigate the following issues: 
\begin{itemize}
\item Effects of missing test results on X\% of datasets (alternative MTD);
\item Effects of missing X\% of test results of algorithms on each dataset (alternative MTA).
\end{itemize}

\noindent
If the performance drop of alternative MTA were not too different from the drop of alternative MTD, than we could conclude that 
X\% of omissions is not unduly degrading the performance and 
hence the method of average ranking is relatively robust. 
Each of these alternatives is discussed in more detail below.

\emph{Missing all test results on some datasets (alternative MTD):}
This strategy involves randomly omitting all test results on a given proportion of datasets from our meta-dataset. An example of this scenario is depicted in Table \ref{tab:strategy:mtd}. In this example the test results on datasets $D_2$ and $D_5$ are completely missing. The aim is to show how much the average ranking degrades due these missing results. 

\begin{table}[htbp] 
\floatconts 
{tab:strategy:mtd}% label 
{\caption{Missing test results on certain percentage of datasets (MTD)}}% caption command 
{% 
\begin{tabular}{|l|l|l|l|l|l|l|} 
  \hline          
  $Algorithms$ & $D_1$ & $D_2$ & $D_3$ &$D_4$& $D_5$ &$D_6$ \\ 
  \hline 
  $a_1$ & $0.85$ &  \cellcolor[gray]{0.5}$$ & $0.77$ &  $0.98$ & \cellcolor[gray]{0.5}$$ &  $0.82$  \\
  \hline 
  $a_2$ & $0.95$ &  \cellcolor[gray]{0.5}$$ & $0.67$ &  $0.68$ & \cellcolor[gray]{0.5}$$ &  $0.72$  \\ 
  \hline 
  $a_3$ & $0.63$ &  \cellcolor[gray]{0.5}$$ & $0.55$ &  $0.89$ & \cellcolor[gray]{0.5}$$ &  $0.46$  \\ 
  \hline  
  $a_4$ & $0.45$ &  \cellcolor[gray]{0.5}$$ & $0.34$ &  $0.58$ & \cellcolor[gray]{0.5}$$ &  $0.63$  \\ 
  \hline 
  $a_5$ & $0.78$ &  \cellcolor[gray]{0.5}$$ & $0.61$ &  $0.34$ & \cellcolor[gray]{0.5}$$ &  $0.97$  \\ 
  \hline 
  $a_6$ & $0.67$ &  \cellcolor[gray]{0.5}$$ & $0.70$ &  $0.89$ & \cellcolor[gray]{0.5}$$ &  $0.22$   \\ 
  \hline 
   
\end{tabular}
} 
\end{table}

\emph{Missing some algorithm test results on each dataset (alternative MTA):} 
Here the aim is to drop a certain proportion of test results on each dataset. 
The omissions are simply distributed uniformly across all datasets. 
That is, the probability that the test result of algorithm $a_i$ is missing 
is the same irrespective of which algorithm is chosen. An example of this scenario is depicted in Table \ref{tab:strategy:mta}. 
%%%%%%%%%%%%%%%%%%%%%%%%%%%%%%%%%%%%%%%%%%%%%%%%%%%%%%%%%%%%%%%%%%%%%%%%%%%%%%
\begin{table}[htbp] 
\floatconts 
{tab:strategy:mta}% label 
{\caption{Missing test results on certain percentage of algorithms (MTA)}}% caption command 
{% 
\begin{tabular}{|l|l|l|l|l|l|l|} 
  \hline          
  $Algorithms$ & $D_1$ &$D_2$ & $D_3$ &$D_4$& $D_5$ &$D_6$ \\ 
  \hline 
  $a_1$ & $0.85$ &  $0.77$ & \cellcolor[gray]{0.5}$$ &  $0.98$ & \cellcolor[gray]{0.5}$$ &  $0.82$ \\
  \hline 
  $a_2$ & \cellcolor[gray]{0.5}$$ & $055$ & $0.67$ &  $0.68$ & $0.66$ & \cellcolor[gray]{0.5}$$ \\ 
  \hline 
  $a_3$ & $0.63$ &  \cellcolor[gray]{0.5}$$ & $0.55$ &  $0.89$ & \cellcolor[gray]{0.5}$$ &  $0.46$  \\ 
  \hline  
  $a_4$ & $0.45$ &  $0.52$ & $0.34$ &  \cellcolor[gray]{0.5}$$ & $0.44$ &  $0.63$  \\ 
  \hline 
  $a_5$ & $0.78$ &  $0.87$ & $0.61$ &  $0.34$ & $0.42$ &  \cellcolor[gray]{0.5}$$  \\ 
  \hline 
  $a_6$ & \cellcolor[gray]{0.5}$$ &  $0.99$ & \cellcolor[gray]{0.5}$$ &  $0.89$ & \cellcolor[gray]{0.5} &  $0.22$   \\ 
  \hline 
   
\end{tabular}
} 
\end{table}
%%%%%%%%%%%%%%%%%%%%%%%%%%%%%%%%%%%%%%%%%%%%%%%%%%%%%%%%%%%%%%%%%%%%%%%%%%%%%%%%%%
The proportion of datasets is a parameter of the method. Here we use the values shown in Table \ref{tab:omissions}. The meta-dataset used in this study is described further on in Section 3.2.
This dataset was used to obtain a new one in which some datasets would be chosen at random and all test results obliterated.
The resulting dataset was used to construct the average ranking.
Each ranking was then used to construct a \emph{loss-time curve} described further on
(in Section \ref{sec:results-effects-omissions}). 
The whole process was repeated 10 times.
This way we would obtain 10 loss-time curves, 
which would be aggregated into a single loss-time curve. 
Our aim is to upgrade the average ranking method to be able to deal with incomplete rankings.  
The enhanced method is described in the next section.
%%%%%%%%%%%%%%%%%%%%%%%%%%%%%%%%%%%%%%%%%%%%%%%%%%%%%%%%%%%%%%%%%%%%%%%%%%%%%%%%%%%%%%%%%%%%
\begin{table}[htbp] 
\floatconts 
{tab:omissions}% label 
{\caption{Percentages of omissions and the numbers of datasets and algorithms used}}% caption command 
{% 
\begin{tabular}{|l|l|l|l|l|l|l|l|} 
 \hline                  
  Omissions \% &  $0$ &  $5$  & $10$  & $20$  & $50$  & $90$  & $95$\\ 
  \hline 
  No of datasets used & $38$ &  $36$  & $34$  & $30$  & $19$  & $4$  & $2$\\ 
  \hline
   No of tests per dataset & $53$ &  $50$  & $48$  & $43$  & $26$  & $5$  & $3$\\ 
  \hline  
\end{tabular} 
} 
\end{table}
%%%%%%%%%%%%%%%%%%%%%%%%%%%%%%%%%%%%%%%%%%%%%%%%%%%%%%%%%%%%%%%%%%%%%%%%%%%%%%%%%%%%%

\subsection{Aggregation Method for Incomplete Rankings (AR-MTA) } 
%(AR_{MTA})
\label{aggregation-method}	
Before describing the method for the calculation of the average ranking that can process incomplete rankings, let us consider a motivating example (see Table \ref{tab:subtabex}), 
illustrating why we cannot simply use the usual average ranking method (\citet{Lin2010}), 
often used in comparative studies in machine learning literature.
%%%%%%%%%%%%%%%%%%%%%%%%%%%%%%%%%%%%%%%%%%%%%%%%%%%%%%%%%%%%%%%%%%%%%%%%%%%%%%%%%%%%%%%%%%
\begin{table}[htbp]
\floatconts
 {tab:subtabex}
 {\caption{An example of two rankings  $R_1$ and  $R_2$ and the aggregated ranking $R^A$}}
 {%
   \subtable{%
     \label{tab:ab}%
     \begin{tabular}{|l|l|}
     \hline          
     $R_1$ & Rank   \\ 
     \hline 
     $a_1$ & $1$   \\
     \hline 
     $a_3$ & $2$    \\ 
     \hline 
     $a_4$ & $3$    \\ 
     \hline 
     $a_2$ & $4$   \\ 
     \hline 
     $a_6$ & $5$    \\ 
     \hline 
     $a_5$ & $6$    \\ 
     \hline 
     \end{tabular}
   }\qquad
   \subtable{%
     \label{tab:cd}%
     \begin{tabular}{|l|l|}
     \hline            
     $R_2$ & Rank   \\ 
     \hline 
     $a_2$ & $1$   \\
     \hline 
     $a_1$ & $2$    \\ 
     \hline 
     \end{tabular}
   }\qquad
   \subtable{%
     \label{tab:ef}%
     \begin{tabular}{|l|l|l|}
    \hline          
    $R^A$ & Rank & Weight  \\ 
    \hline 
    $a_1$ & $1.67$ &  $1.2$   \\
    \hline 
    $a_3$ & $2$ &  $1$   \\ 
    \hline 
    $a_4$ & $3$ &  $1$   \\ 
    \hline 
    $a_2$ & $3.5$ &  $1.2$   \\ 
    \hline 
    $a_6$ & $5$ &  $1$   \\ 
    \hline 
    $a_5$ & $6$ &  $1$   \\ 
    \hline 
    \end{tabular}
    }
  }
\end{table}
%
% %%%%%%%%%%%%%%%%%%%%%%%%%%%%%%%%%%%%%%%%%%%%%%%%%%%%%%%%%%%%%%%%%%%% 
%  
\noindent
Let us compare the rankings $R_1$ (Table \ref{tab:subtabex}a) and $R_2$ (Table \ref{tab:subtabex}b).
We note that algorithm $a_2$ is ranked 4 in ranking  $R_1$, 
but has rank 1 in ranking  $R_2$. If we used the usual method, the final ranking of $a_2$ would be the mean of the two ranks, i.e. (4+1)/2=2.5. 
This seems intuitively not right, as the information in ranking  $R_2$ is incomplete. 
If we carry out just one test and obtain ranking $R_2$ as a result, 
this information is obviously inferior to having conducted more tests 
leading to ranking  $R_1$. 
So these observations suggest that the number of tests should be taken into account 
to set a weight to characterize the individual elements of the ranking. 

In our method the weight is calculated using the expression $(N-1)/(Nmax-1)$, where $N$ represents the number of filled-in elements in the ranking and $Nmax$ the maximum number of elements that could be filled-in. The number $N-1$ represents the number of non-transitive relations between any element in the ranking to any other element in the ranking.
So, for instance, in ranking  $R_1$ $N=6$ and $Nmax=6$. Therefore, the weight of each element in the ranking is $5/5=1$. We note that $N-1$ (i.e. 5), 
represents the number of non-transitive relations in the ranking, namely $a_1>a_3$, $a_3>a_4$,~..~, $a_6>a_5$. Here $a_i>a_j$ is used to indicate that $a_i$ is preferred to $a_j$.

Let us consider the incomplete ranking  $R_2$. Suppose we know a priori that the ranking could include 6 elements and so $Nmax=6$, as in the previous case. 
Then the weight of each element will be $(N-1)/(Nmax-1) = 1/5=0.2$. 
The notion of \emph{weight} captures the fact that ranking  $R_2$ provides less information than ranking  $R_1$.  We need this concept in the process of calculating the average ranking. 

%\subsubsection{Overview of the aggregation method for incomplete rankings}
Our upgraded version of the aggregation method for incomplete rankings involves the initialization step, which consists of reading-in the first ranking and initializing the average ranking $R^A$. Then in each subsequent step a new ranking is read-in and aggregated with the average ranking, producing a new average ranking. 
The aggregation is done by going through all elements in the ranking, one by one. If the element appears in both the aggregated ranking and the read-in ranking, its rank is recalculated as a weighted average of the two ranks:
\begin{equation}
r_i^A := r_i^A * w_i^A /(w_i^A+w_i^j)  + r_i^j * w_i^j/(w_i^A+w_i^j) 
\end{equation}

\noindent
where $r_i^A$ represents the rank of element $i$ in the aggregated ranking and $r_i^j$ the rank of the element in the ranking $j$ that is being processed
and $w_i^A$ and $w_i^j$ represent the corresponding weights. 
The weight is updated as follows:
$w_i^A := (w_i^A+w_i^j)$.
If the element appears in the aggregated ranking, but not in the new read-in ranking, the new aggregated ranking is made equal to the previous one. The weight of the element is also kept unchanged.
 
Suppose the aim is to aggregate ranking  $R_1$ and  $R_2$ shown before. The new rank of $a_2$ will be $r_2^A$ = 4 * 1/1.2 + 1*0.2/1.2 = 3.5. The weight will be $w_2^A$ = 1 + 0.2 = 1.2.
The final aggregated ranking of rankings  $R_1$ and  $R_2$ is shown in Table \ref{tab:subtabex}c.

\subsection{Results on the Effects of Omissions in the Meta-Dataset}
\label{sec:results-effects-omissions}

The data used in the experiments involves the meta-dataset constructed from 
evaluation results retrieved from OpenML~(\citet{Vanschoren2014}), 
a collaborative science platform for machine learning. 
This dataset contains the results of $53$ parameterized classification algorithms from the Weka workbench~(\citet{Hall2009}) on $39$ datasets\footnote{Full details: \url{http://www.openml.org/project/tag/ActiveTestingSamples/u/1}}. 
In the leave-one-out mode, $38$ datasets are used to generate the model (e.g. average ranking), while the dataset left out is used for evaluation. 

\begin{figure}[ht!]
  \begin{center}
    \includegraphics[width=.50\textwidth]{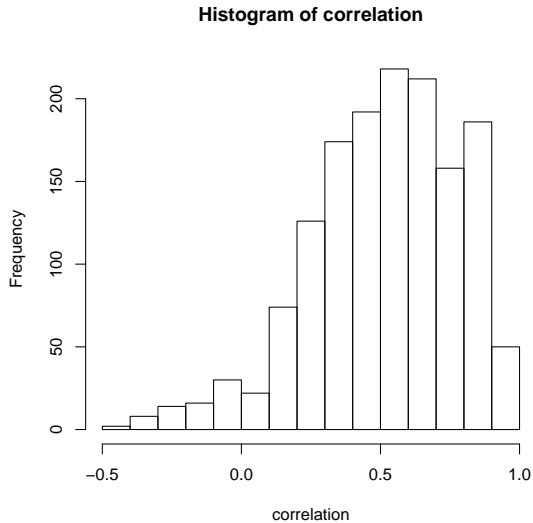}
  \end{center}
  \caption{Spearman correlation between rankings of pairs of datasets.}
  \label{fig:compar-rankings}
\end{figure}

%\noindent
\emph{Characterization of our Meta-Dataset:}
\noindent
We are interested to analyze rankings of classification algorithms 
on different datasets and in particular how these differ for pairs of datasets,
using Spearman correlation coefficient.
Fig.\ref{fig:compar-rankings} shows a histogram characterizing the meta-dataset used.
The histogram is accompanied by \emph{expected value}, \emph{standard deviation} and \emph{coefficient of variation} calculated as the ratio of standard deviation to the expected value (mean) ~(\citet{Witten2005}). These measures are shown in Table \ref{fig:measures-compar-rankings}.

% some basic characteristic of our meta-dataset
%
\begin{table}[htbp] 
\floatconts 
{tab:characteristics}% label 
{\caption{Measures characterizing the histogram of correlations in Fig. \ref{fig:compar-rankings} }}% caption command
{ 
\label{fig:measures-compar-rankings} 
\begin{tabular}{|l|l|l|l|} 
 \hline                  
  Measure \% &  Expected Value & Standard Deviation & Coefficient of Variation \\ 
  \hline 
  Value & $0.5134$ &  $0.2663$  & $51.86\%$ \\ 
  \hline   
\end{tabular} 
}
\end{table}
 
\emph{Results:} 
The aim is to investigate how certain omissions in the meta-datasets affect the performance. 
The results are presented in the form of loss-time curve (\citet{Rijn2015})
which show how \emph{performance loss} depends on time. 
The \emph{loss} is calculated as the difference between the performance 
of the best algorithm identified so far when following the ranking 
to the ideal choice.
Each loss-time curve can be characterized by a number representing 
the \emph{mean loss} in a given interval. 
We want this \emph{mean interval loss} (MIL) 
to be as low as possible. 
This characteristic is similar to AUC, but there is an important difference.  
When talking about AUCs, the x-axis values spans between 0 and 1. 
Our loss-time curves span between some $T_{min}$ and $T_{max}$ 
and both values depend on the user. 
Typically the user searching for a suitable algorithm would not worry about quite short times. 
In the experiments here we have set $T_{min}$ to 10 seconds. 
The value of $T_{max}$ was set to $10^4$ seconds, i.e. about 2.78 hours. 

\begin{figure}[ht!]
  \begin{center}
    \includegraphics[width=.53\textwidth]{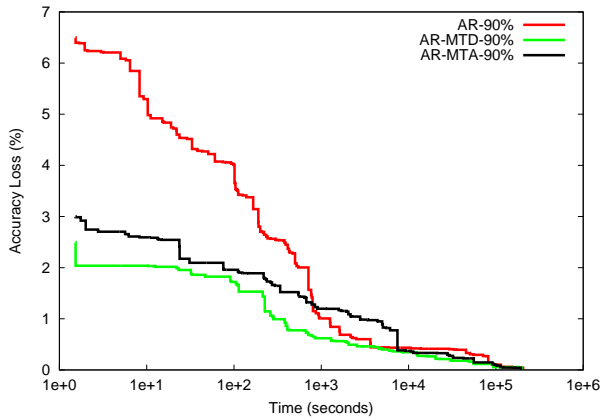}
  \end{center}
  \caption{Comparison of \emph{AR-MTA} with \emph{AR-MTD} and the baseline method \emph{AR} for 90\% of omissions.}
  \label{fig:loss-cuves-missing}
\end{figure}

Table \ref{tab:MIL-two-strategies} presents the results for both alternative \emph{AR-MTD} and \emph{AR-MTA} in terms of mean interval loss (MIL).
The values for ordinary average ranking method, \emph{AR}, are also shown, 
as this method serves as a baseline.
Fig.\ref{fig:loss-cuves-missing} shows the loss-time curves 
for the alternatives \emph{AR-MTD} and \emph{AR-MTA} when the number of omissions is 90\%.
The values of \emph{AR} relative to 90\% omissions are also shown for comparison.
Not all loss-time curves are shown, as the figure would be rather cluttered.

%\begin{table}[htbp] 
%\floatconts 
%{tab:MIL-two-strategies}% label 
%{\caption{Mean interval loss values for different percentage of omissions for the two strategies}}% caption command 
%{% 
%\begin{tabular}{|l|l|l|l|l|l|l|l|l|} 
% \hline       
% \backslashbox{Method}{Omission\%}&  & $0\%$ & $5\%$  &  $10\%$ & $20\%$  &  $50\%$  & $90\%$  &  $95\%$ \\ 
% \hline 
% Baseline  & MIL & $1.4066$ & $2.330$ & $2.477$ &  $2.825$ & $2.531$   &  $2.324$ & $2.384$  \\
%  \hline 
% Alternative 1  & MIL & $1.4066$ & $1.129$ & $0.812$ &  $0.937$ & $1.046$   &  $0.981$ & $1.387$  \\
% \hline 
% Alternative 2 &  MIL & $1.4066$ &  $0.881$ & $0.971$ & $0.972$ &  $1.031$ &  $1.473$ & $1.799$  \\ 
%  \hline 
%%  \ $MIL_2$/$MIL_1$ &  & & $0.780$  & $1.196$ &  $1.037$  & $0.986$ &  $1.501$  & $1.297$ \\ 
%%  \hline 
%\end{tabular} 
%} 
%\end{table} 

\begin{table}[htbp] 
\floatconts 
{tab:MIL-two-strategies}% label 
{\caption{Mean interval loss (MIL) values for different percentage of omissions}}% caption command 
{% 
\begin{tabular}{|l|l|l|l|l|l|l|l|l|} 
 \hline       
 \backslashbox{Method}{Omission\%} & $0\%$ & $5\%$  &  $10\%$ & $20\%$  &  $50\%$  & $90\%$  &  $95\%$ \\ 
 \hline 
  AR  & $1.213$ & $2.306$ & $2.435$ &  $2.785$ & $2.507$   &  $2.302$ & $2.360$  \\
  \hline 
  AR-MTD & $1.213$ & $1.129$ & $0.812$ &  $0.937$ & $1.046$   &  $0.981$ & $1.387$  \\
 \hline 
  AR-MTA & $1.213$ &  $0.881$ & $0.971$ & $0.972$ &  $1.031$ &  $1.473$ & $1.799$  \\ 
  \hline 
%  \ $MIL_2$/$MIL_1$ &  & & $0.780$  & $1.196$ &  $1.037$  & $0.986$ &  $1.501$  & $1.297$ \\ 
%  \hline 
\end{tabular} 
} 
\end{table} 
%%%%%%%%%%%%%%%%%%%%%%%%%%%%%%%%%%%%%%%%%%%%%%%%%%%%%%%%
 
\noindent
Our results show that the alternative \emph{AR-MTA} 
achieves far better results than the baseline method \emph{AR}. 
We note also that although the proposed method \emph{AR-MTA} is worse than 
the \emph{AR-MTD} counterpart, the difference is negligible 
for many different values of omissions in the range from 5\% till 50\%.
Only when we get to rather extreme values such as 90\%, the difference is noticeable. 
But even in this case the curve of \emph{AR-MTA} follows the curve of \emph{AR-MTD} much more closely than \emph{AR}.
These results indicate that the proposed average ranking method is relatively robust to omissions.

\emph{Future work:}
As the incomplete meta-dataset does not affect much 
the final ranking and the corresponding loss, 
this could be explored in future design of experiments, 
when gathering the test results.
We could investigate approaches that permit to consider also the costs (time) of off-line tests. Their cost (time) could be set to some fraction of the cost of on-line test (i.e. tests on a new dataset), but not really ignored altogether. \\
\\
\noindent
\textbf{Acknowledgements}\\
This work is supported by TETFund 2012 Intervention for Kano University of Science and Technology, Wudil, Kano State, Nigeria for PhD Overseas Training from the Federal Government of Nigeria.

The authors  also acknowledge the support of project \emph{NanoSTIMA: Macro-to-Nano Human Sensing: Towards Integrated Multimodal Health Monitoring and Analytics/NORTE-01-0145-FEDER-000016}, which is financed by the North Portugal Regional Operational Programme (NORTE 2020), under the PORTUGAL 2020 Partnership Agreement, and through the European Regional Development Fund (ERDF).

% Bibliography starts here

\bibliography{jmlrwcp-sample}

\end{document}